%% file: root.tex
\documentclass[letterpaper, 10 pt, conference]{ieeeconf}
\input{packages}

\input{macros}
\input{acronyms}
\IEEEoverridecommandlockouts          
\renewcommand{\baselinestretch}{0.98}
\newcommand{\method}{CoPa}
\definecolor{myorange}{HTML}{FD6238}
\definecolor{myblue}{HTML}{0065A2}
\definecolor{mygreen}{HTML}{649A40}

\title{\LARGE \bf
\method: General Robotic Manipulation through \\ Spatial Constraints of Parts with Foundation Models
\vspace{-0.5em}
\author{Haoxu Huang$^{2,3,4*}$, Fanqi Lin$^{1,2,4*}$, Yingdong Hu$^{1,2,4}$, Shengjie Wang$^{1,2,4}$, Yang Gao$^{1,2,4}$%
\thanks{$^{*}$ The first two authors contributed equally.}%
\thanks{$^{1}$ Institute of Interdisciplinary Information Sciences, Tsinghua University.}%
\thanks{$^{2}$ Shanghai Qi Zhi Institute.}%
\thanks{$^{3}$ Shanghai Jiao Tong University.}%
\thanks{$^{4}$ Shanghai Artificial Intelligence Laboratory.}
}}

\begin{document}

\newcommand{\insertteaser}{
    \vspace{1em}
    \includegraphics[width=\linewidth]{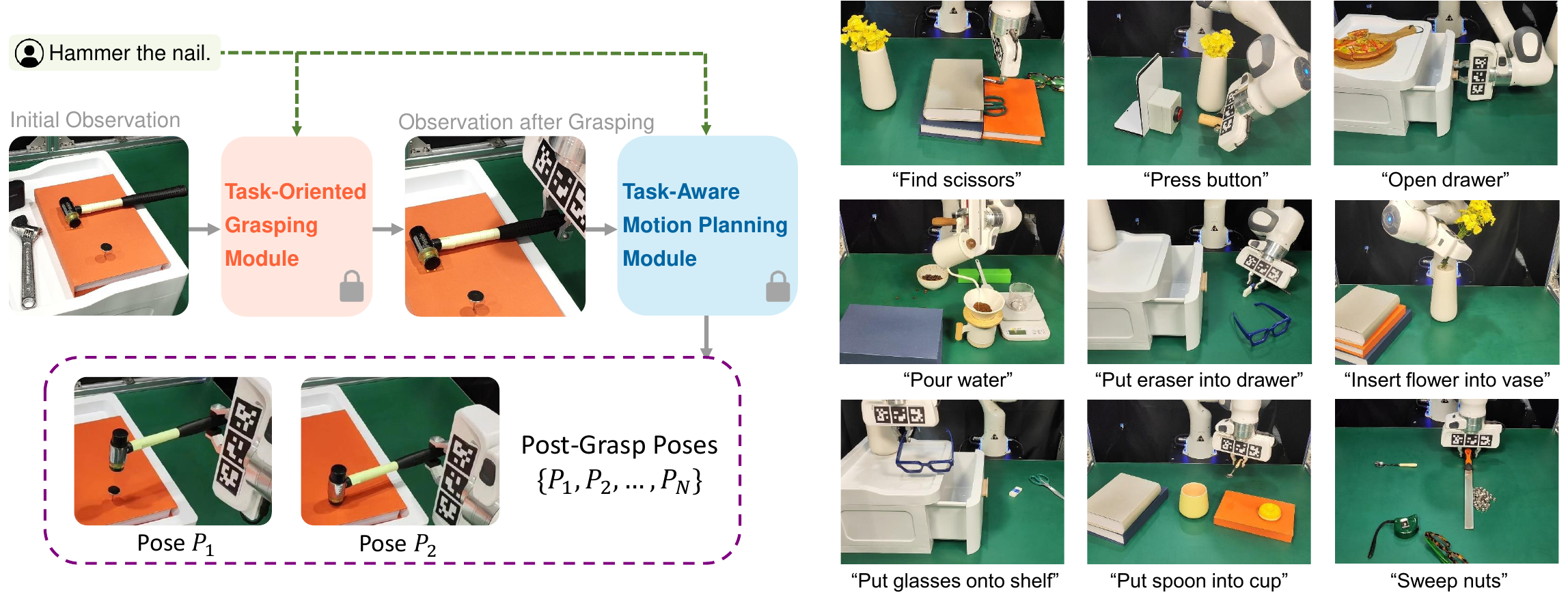}
    \captionof{figure}{\textbf{Overview.} We present \method, a novel framework that utilizes common sense knowledge embedded within VLMs for robotic low-level control.
        \textbf{Left.} Our pipeline. Given an instruction and scene observation, \method~first generates a grasp pose through \textcolor{myorange}{Task-Oriented Grasping Module} (detailed in Fig. \ref{fig: method-grasp}).
        Subsequently, a \textcolor{myblue}{Task-Aware Motion Planning Module} (detailed in Fig. \ref{fig: method-bahavior}) is utilized to obtain post-grasp poses.
        \textbf{Right.} Examples of real-world experiments.
        Boasting a fine-grained physical understanding of scenes,
        \method~can generalize to open-world scenarios, handling open-set instructions and objects with minimal prompt engineering and without the need for additional training.}
    \label{fig: method-overview}
}

\makeatletter
\apptocmd{\@maketitle}{\centering\insertteaser}{}{}
\makeatother

\maketitle
\setcounter{figure}{1}


\thispagestyle{empty}
\pagestyle{empty}

\input{contents/abstract}
\input{contents/intro}

\input{contents/related_work}

\input{contents/method}
\input{contents/experiments}
\input{contents/conclusions}

\bibliographystyle{IEEEtran}
\bibliography{root}

\input{contents/appendix}

\end{document}

%% file: packages.tex
\usepackage[binary-units=true]{siunitx}
\usepackage{times}
\usepackage{amsmath,amssymb}
\usepackage[nolist,nohyperlinks]{acronym}
\usepackage{accents}
\usepackage{graphicx}
\usepackage{bm}
\usepackage{bbm}
\let\labelindent\relax
\usepackage[inline]{enumitem}
\usepackage{array}
\usepackage{multicol}
\usepackage{multirow}
\usepackage{tabulary}
\usepackage{booktabs}
\usepackage{subfig}
\usepackage{todonotes}
\usepackage{cite}
\usepackage{algorithm}
\usepackage{algpseudocode}
\usepackage[bookmarks=true]{hyperref}
\usepackage{xcolor}

%% file: acronyms.tex
\begin{acronym}
	\acro{CoM}{Center of Mass}
	\acro{DFS}{Depth First Search}
	\acro{CITO}{Contact-Implicit Trajectory Optimization}
	\acro{MIP}{Mixed-Integer Programming}
	\acro{MIQP}{Mixed-Integer Quadratic Programming}
	\acro{RNN}{Recurrent Neural Network}
	\acro{MLP}{Multilayer Perceptron}
	\acro{S}[\emph{S}]{\textbf{sliding}}
	\acro{SC}[\emph{SC}]{\textbf{sliding with curvature}}
	\acro{R}[\emph{R}]{\textbf{rotating}}
	\acro{P}[\emph{P}]{\textbf{pivoting}}
	\acro{L}[\emph{L}]{\textbf{lifting}}
	\acro{MCTS}{Monte Carlo Tree Search}
	\acro{PVMCTS}{Policy-Value Monte Carlo Tree Search}
	\acro{MDP}{Markov-decision Process}
	\acrodefplural{MDP}{Markov-decision Processes}
	\acro{ADMM}{Alternating Direction Method of Multipliers}
	\acro{QP}{Quadratic Program}
	\acro{DoF}{Degree of Freedom}
	\acrodefplural{DoF}{Degrees of Freedom}%
	\acro{leg}[\emph{L}]{\emph{legible}}		
\end{acronym}

%% file: contents/abstract.tex
\begin{abstract}
Foundation models pre-trained on web-scale data are shown to encapsulate extensive world knowledge beneficial for robotic manipulation in the form of task planning. 
However, the actual physical implementation of these plans often relies on task-specific learning methods, which require significant data collection and struggle with generalizability.
In this work, we introduce Robotic Manipulation through
Spatial \underline{Co}nstraints of \underline{Pa}rts (\textbf{\method}), a novel framework that leverages the common sense knowledge embedded within foundation models to generate a sequence of 6-DoF end-effector poses for open-world robotic manipulation.
Specifically, we decompose the manipulation process into two phases: task-oriented grasping and task-aware motion planning. 
In the task-oriented grasping phase, we employ foundation vision-language models (VLMs) to select the object’s grasping part through a novel coarse-to-fine grounding mechanism.
During the task-aware motion planning phase, VLMs are utilized again to identify the spatial geometry constraints of task-relevant object parts, which are then used to derive post-grasp poses.
We also demonstrate how \method~can be seamlessly integrated with existing robotic planning algorithms to accomplish complex, long-horizon tasks.
Our comprehensive real-world experiments show that \method~possesses a fine-grained physical understanding of scenes, capable of handling open-set instructions and objects with minimal prompt engineering and without additional training.
Project page: \href{https://copa-2024.github.io/}{copa-2024.github.io}

\end{abstract}

%% file: contents/intro.tex
\section{INTRODUCTION}


Developing a general-purpose robot necessitates effective approaches in two critical areas: \textit{(i)} high-level task planning, which determines what to do next, and \textit{(ii)} low-level robotic control, focusing on the precise actuation of joints~\cite{cambon2009hybrid,kaelbling2011hierarchical}.
The emergence of high-capacity foundation models~\cite{bommasani2021opportunities,achiam2023gpt}, pre-trained on extensive web-scale datasets, has inspired a surge of recent research efforts aimed at integrating these models into robotics~\cite{hu2023toward,firoozi2023foundation}. 
Nonetheless, these methods generally address only the ``higher level'' aspects of task planning~\cite{ahn2022can,huang2023grounded,huang2022inner,hu2023look}. In contrast, the prevailing approach for low-level control continues to revolve around crafting task-specific policies via diverse learning methods~\cite{hussein2017imitation,sutton2018reinforcement}. Such policies, however, are brittle and prone to failure when encountering unseen scenarios~\cite{xie2023decomposing}. 
Even the largest robotics models struggle outside environments they have previously encountered~\cite{brohan2022rt,brohan2023rt}.

The question then arises: what makes generalizable low-level robotic control so hard? We attempt to answer this question through the lens of human object manipulation. 
For instance, when an individual is tasked with hammering a nail, regardless of their familiarity with the specific hammer, they intuitively grasp it by the handle (instead of the head), adjust its orientation so the striking surface aligns with the nail, and then execute the strike.
This process underscores the importance of a fine-grained understanding of the physical properties of task-related objects, or more broadly, the extensive common sense knowledge of the world that facilitates generalizable object manipulation. 
Some pioneering works~\cite{huang2023voxposer,liang2023code} have sought to leverage the rich semantic knowledge of Internet-scale foundation models to enhance low-level robotic control. 
Yet, these approaches are heavily dependent on intricate prompt engineering and suffer from a fundamental limitation: a \textit{coarse} understanding of the scene, leading to failures in tasks requiring \textit{fine-grained} physical understanding. Such a detailed understanding is essential for nearly all real-world robotic tasks of interest.


To endow robots with fine-grained physical understanding, we propose Robotic Manipulation through
Spatial \textbf{Co}nstraints of   \textbf{Pa}rts (\textbf{\method}), a novel framework that incorporates common sense knowledge embedded within foundation vision-language models (VLMs), such as GPT-4V, into the robotic manipulation tasks.
We observe that most manipulation tasks require a part-level, fine-grained physical understanding of objects within the scene.
Hence, we design a coarse-to-fine grounding module to identify task-relevant parts.
Then, to leverage VLMs for aiding the robotic low-level control, it is necessary to design an interface that not only allows VLMs to reason in the form of language but also facilitates robot's object manipulation.
Therefore, we propose utilizing \textit{spatial constraints} as a bridge between VLMs and robots.
Specifically, we utilize VLMs to generate the spatial constraints that task-relevant parts must meet to accomplish the task, and then employ a solver to determine the robot's poses based on these constraints.
Finally, to ensure the precise execution of the robot's actions, transitions between adjacent poses are achieved through traditional motion planning methods.

We demonstrate that \method~is capable of completing everyday manipulation tasks with a high success rate through extensive real-world experiments. 
Attributed to the innovative design of coarse-to-fine grounding and constraint generation module, \method~possesses a profound physical understanding of the environment and can generate precise 6-Dof poses to complete complex manipulation tasks, significantly surpassing a strong baseline VoxPoser~\cite{huang2023voxposer}.

Our contributions are summarized as follows:
\begin{itemize}
    \item We propose \method, a novel framework that utilizes the common sense knowledge of VLMs for low-level robotic control, which can handle open-set instructions and objects with minimal prompt engineering and without additional training.
    \item Through extensive real-world experiments, \method~is demonstrated to possess the capability to complete manipulation tasks that require a fine-grained understanding of physical properties of task-relevant objects, significantly surpassing baselines.
    \item We show that \method~can be seamlessly integrated with high-level planning methods to accomplish complex, long-horizon tasks (e.g. make pour-over coffee and set up romantic table).
\end{itemize}


%% file: contents/related_work.tex
\section{RELATED WORK}
\label{sec:related-work}


\noindent \textbf{Learning for Robotic Manipulation.}
Manipulation is a critical and challenging aspect in the robotic field. 
Numerous studies harness imitation learning (IL) from expert demonstrations to acquire manipulation skills~\cite{brohan2022rt,shridhar2022cliport, shridhar2023perceiver,zhang2023universal,brohan2023rt,padalkar2023open,team2023octo}. 
Despite IL's conceptual simplicity and its notable success across a broad spectrum of real-world tasks, it struggles with out-of-distribution samples and demands considerable effort in collecting expert data.
Reinforcement learning (RL)~\cite{sutton2018reinforcement} emerges as another principal approach~\cite{liu2023imitation, ye2023foundation, ahn2022can, akkaya2019solving, levine2016end}, enabling robots to develop manipulation skills via trial-and-error interactions with their environment. 
However, RL's sample inefficiency limits its applicability in real-world settings, leading most robotic systems to rely on sim-to-real transfers~\cite{xu2023roboninja, matas2018sim, jeong2020self}. 
Nonetheless, sim-to-real approaches necessitate the construction of specific simulators and confront the sim-to-real gap. 
Furthermore, policies learned via these end-to-end learning methods often lack generalization to new tasks. 
In contrast, by leveraging foundation models' common sense knowledge, our \method~can generalize to open-world scenarios without additional training.

\noindent \textbf{Foundation Models For Robotics.}
In recent years, foundation models have dramatically transformed the landscape of robotics~\cite{hu2023toward}. 
Many works employ vision models, pre-trained on large-scale image datasets, to generate visual representations for visuomotor control tasks~\cite{parisi2022unsurprising, nair2022r3m, radosavovic2023real, xiao2022masked, zhang2023universal, majumdar2023we}.
Some other studies utilize foundation models for reward specification in reinforcement learning~\cite{ma2022vip, ma2023eureka, alakuijala2023learning, mahmoudieh2022zero, cui2022can, ma2023liv}.
Furthermore, numerous studies have leveraged foundation models for robotic high-level planning, achieving remarkable success~\cite{hu2023look, ahn2022can, huang2023grounded, singh2023progprompt, gao2023physically, ding2023task, huang2022language, lin2023text2motion, liu2023llm+, ren2023robots, song2023llm, wu2023tidybot}.
There is also a body of works that employs foundation models for low-level control~\cite{brohan2022rt, brohan2023rt, padalkar2023open, team2023octo}. Some works fine-tune vision-language models (VLMs) to directly output robot actions. However, such fine-tuning approaches require extensive amounts of expert data.
To address this issue, Code as Policies~\cite{liang2023code} applies large language models (LLMs) to write code to control robots, and VoxPoser~\cite{huang2023voxposer} generates robot trajectories by producing value maps based on foundation models. Nevertheless, these methods rely on complex prompt engineering and possess only a coarse understanding of the scene.
In stark contrast, benefiting from the rational use of common sense knowledge within VLMs, our method exhibits a fine-grained understanding of scenarios and generalizes to open-world scenarios without additional training, requiring only minimal prompt engineering.

%% file: contents/method.tex
\section{METHOD}
\label{sec:method}

\begin{figure*}[t]
    \centering
    \includegraphics[width=\textwidth]{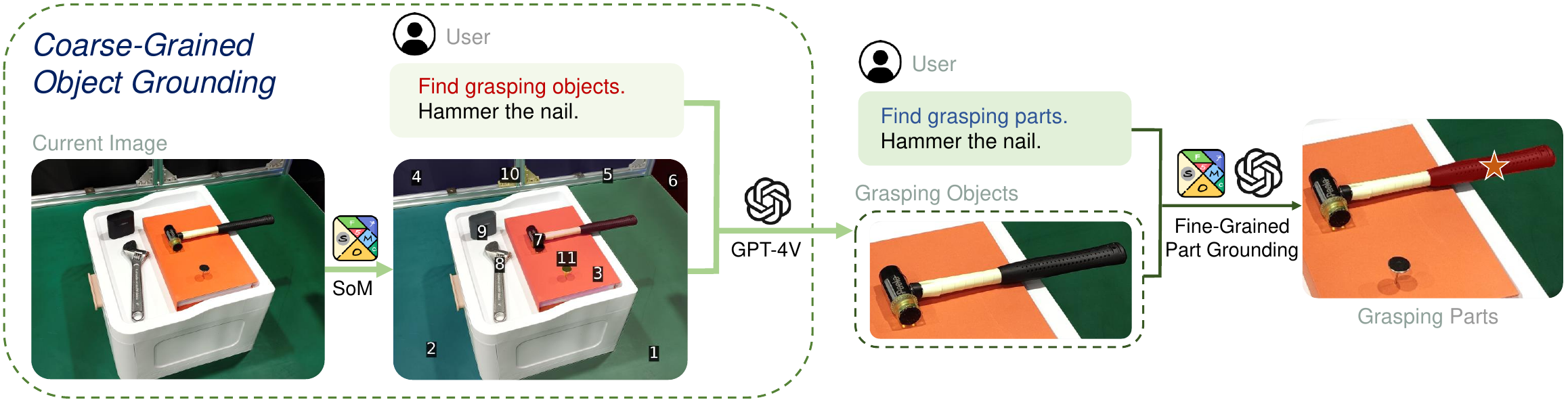}
    \caption{\textcolor{mygreen}{Grounding Module}. This module is utilized to identify the \textcolor{myorange}{grasping part for task-oriented grasping} or \textcolor{myblue}{task-relevant parts for task-aware motion planning}.
    The grounding process is divided into two stages: coarse-grained object grounding and fine-grained part grounding.
    Specifically, we first segment and label objects within the scene using SoM.
    Then, in conjunction with the instruction, we employ GPT-4V to select the \textcolor{myorange}{grasping}/\textcolor{myblue}{task-relevant} objects.
    Finally, similar fine-grained part grounding is applied to locate the specific \textcolor{myorange}{grasping}/\textcolor{myblue}{task-relevant} parts.
    } 
    \label{fig: method-grounding}
\end{figure*}

In this section, we first introduce the formulation of manipulation tasks in Section~\ref{formulate}.
Then, we describe two critical components within our framework — the task-oriented grasping in Section~\ref{grasp} and the task-aware motion planning in Section~\ref{behavior}.

\subsection{Promblem Formulation}
\label{formulate}

Most manipulation tasks can be decomposed into two phases: the initial grasp of the object and the subsequent motion required to complete the task. 
For example, opening a drawer involves grasping the handle and pulling it in a straight line, while picking up a water glass requires first seizing the glass and then lifting it. 
Motivated by this observation, we structure our approach into two modules: \textbf{task-oriented grasping} and \textbf{task-aware motion planning}. 
Additionally, we posit that the execution of robotic tasks essentially entails generating a series of target poses for the robot’s end-effector. The transition between adjacent target poses can be achieved through motion planning.


Given a language instruction $l$ and the initial scene observation $O_0$ (RGB-D images), 
our objective in the task-oriented grasping module is to generate the appropriate grasp pose for the specified objects of interest. This process is represented as $P_0 = f(l, O_0)$. 
We denote the observation after the robot reaches $P_0$ as $O_1$.
For the task-aware motion planning module, our goal is to derive a sequence of post-grasp poses, expressed as 
$g(l, O_1) \longrightarrow \{P_1, P_2, ..., P_N\}$,
where $N$ is the total number of poses required to complete the task.
After acquiring the target poses, the robot's end-effector can reach these poses utilizing motion planning algorithms such as RRT*~\cite{karaman2011sampling} and PRM*~\cite{gang2016prm}.

\begin{figure}[t]
    \centering
    \includegraphics[width=0.5\textwidth]{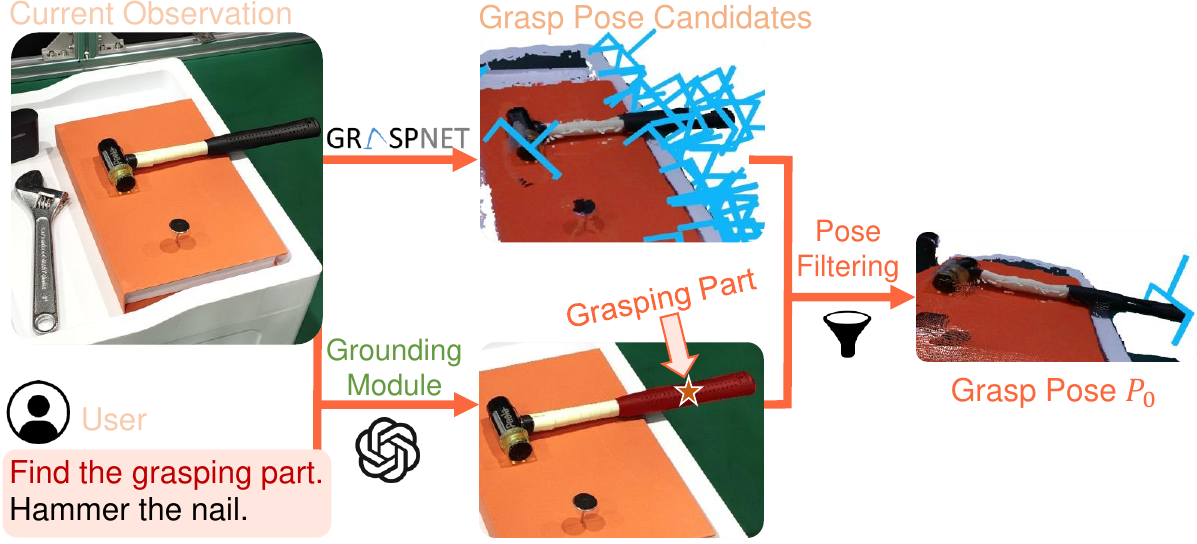}
    \caption{\textcolor{myorange}{Task-Oriented Grasping Module}. This module is employed to generate grasp poses.
    Grasp pose candidates are generated from the scene point cloud using GraspNet.
    Concurrently, given the instruction and the scene image, the grasping part is identified by a \textcolor{mygreen}{grounding module} (detailed in Fig. \ref{fig: method-grounding}).
    Ultimately, the final grasp pose is selected by filtering candidates based on the grasping part mask and GraspNet scores.} 
    \label{fig: method-grasp}
    \vspace{-0.7em}
\end{figure}

\begin{figure*}[t]
    \centering
    \includegraphics[width=\textwidth]{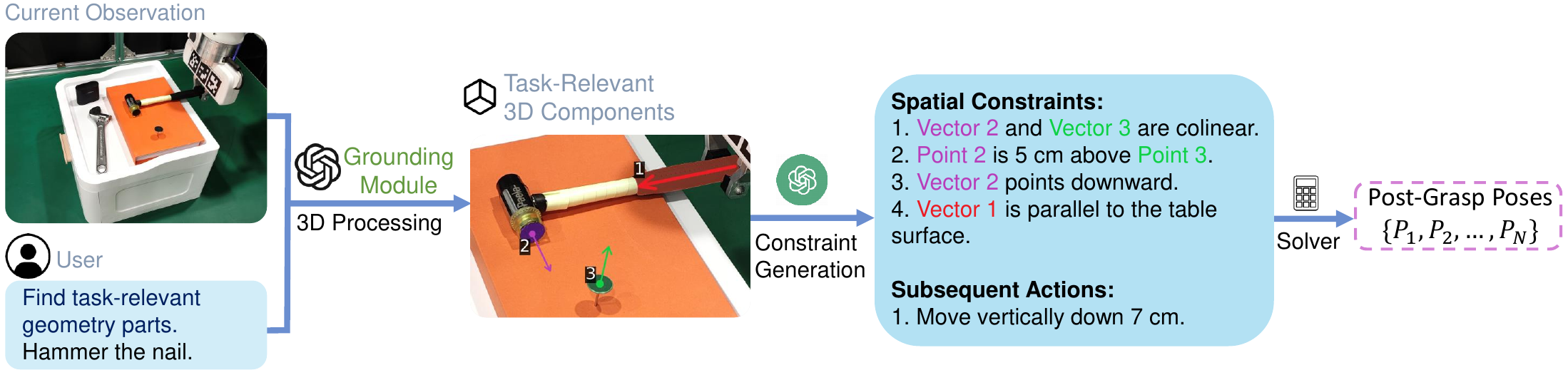}
    \caption{\textcolor{myblue}{Task-Aware Motion Planning Module}. This module is used to obtain a series of post-grasp poses.
    Given the instruction and the current observation, we first employ a \textcolor{mygreen}{grounding module} (detailed in Fig. \ref{fig: method-grounding}) to identify task-relevant parts within the scene.
    Subsequently, these parts are modeled in 3D, and are then projected and annotated onto the scene image.
    Following this, VLMs are utilized to generate spatial constraints for these parts.
    Finally, a solver is applied to calculate the post-grasp poses based on these constraints.} 
    \vspace{-0.7em}
    \label{fig: method-bahavior}
\end{figure*}

\subsection{Task-Oriented Grasping}
\label{grasp}
To generate the task-oriented grasp pose, our approach initially employs a grasping model to produce grasp pose proposals, and filter out the most feasible one through our novel grasping part grounding module. The entire process is depicted in Fig.~\ref{fig: method-grasp}.

\noindent \textbf{Grasp Pose Proposals.}
We leverage a pre-trained grasping model for generating grasp pose proposals. To achieve this, we first convert RGB-D images into point clouds by back-projecting them into 3D space. These point clouds are then input into GraspNet~\cite{fang2020graspnet}, a model trained on a vast dataset comprising over one billion grasp poses. GraspNet outputs 6-DOF grasp candidates, including information on grasp point, width, height, depth, and a ``graspness score,'' which indicates the likelihood of a successful grasp. However, given that GraspNet yields all potential grasps within a scene, it is necessary for us to employ a filtering mechanism that selects the optimal grasp based on the specific task outlined by the language instruction.

\noindent \textbf{Grasping Part Grounding.}
Humans grasp specific parts of an object corresponding to the intended use. For instance, when grasping a knife for cutting, we hold onto the handle rather than the blade; similarly, when picking up glasses, we grasp the frame instead of the lenses. This process essentially represents the application of common sense knowledge by humans. 
To mimic this ability, we utilize vision-language models (VLMs), such as GPT-4V~\cite{gpt4v}, which incorporate vast amounts of common sense knowledge~\cite{hu2023look, geng2023sage}, to identify the appropriate part of an object to grasp.


We employ a two-stage process to ground language instructions to the specific parts of objects intended for grasping: \textit{coarse-grained object grounding} and \textit{fine-grained part grounding}. 
The entire grounding process is shown in Fig. \ref{fig: method-grounding}.
At both stages, we incorporate a recent visual prompting mechanism known as Set-of-Mark (SoM)~\cite{yang2023set}. SoM leverages segmentation models to partition an image into distinct regions, assigning a numeric marker to each, significantly boosting the visual grounding capabilities of VLMs. 
During the \textit{coarse-grained object grounding} phase, SoM is utilized at the object level to detect and label all objects within the scene. Following this, VLMs are tasked with pinpointing the target object for grasping (e.g., a hammer), guided by the user's instructions. 
The selected object is then cropped from the image, upon which \textit{fine-grained part grounding} is applied to determine the specific part of the object to be grasped (e.g., the handle of the hammer). This coarse-to-fine design endows our method with fine-grained physical understanding ability, enabling generalization across complex scenarios. 
Finally, we filter the grasp pose candidates, projecting all the grasp points onto the image and retaining only those within the grasping part mask. From these, the pose with the highest confidence scored by GraspNet is selected as the ultimate grasp pose $P_0$ for execution.


\subsection{Task-Aware Motion Planning}
\label{behavior}

After successfully executing task-oriented grasping, now we aim to obtain a series of post-grasp poses.
We divide this step into three modules: task-relevant part grounding, manipulation constrains generation and target pose planning.
The entire process is shown in Fig. \ref{fig: method-bahavior}.

\noindent \textbf{Task-Relevant Part Grounding.}
Similar to the previous grasp part grounding module, we use \textit{coarse-grained object grounding} and \textit{fine-grained part grounding} to locate task-relevant parts.
Here we need to identify multiple task-relevant parts (e.g. the hammer's striking surface, handle and the nail's surface).
Additionally, we observe that numeric marks on the robotic arm may affect VLMs' selection, so we filter out the masks on the robotic arm (detailed in the Appendix).

\noindent \textbf{Manipulation Constraints Generation.}
During the execution of tasks, task-relevant objects are often subject to various spatial geometric constraints.
For instance, when charging a phone, the charger's connector must be aligned with the charging port;
similarly, when capping a bottle, the lid must be positioned directly above the mouth of the bottle.
These constraints inherently necessitate common sense knowledge, which includes a profound comprehension of the physical properties of objects.
We aim to leverage VLMs to generate spatial geometric constraints for the object manipulated by the robot.

We first model identified task-relevant parts as simple geometric elements.
Specifically, we represent slender parts (e.g. hammer handle) as vectors, while other parts are modeled as surfaces.
For the parts modeled as vectors, we directly draw them on the scene image;
for those modeled as surfaces, we ascertain their center points and normal vectors, which are then projected and marked on the 2D scene image.
The annotated image is used as input for VLMs, which are prompted to generate spatial constraints for these geometric elements. 
We craft a set of descriptions for spatial constraints, such as collinearity between two vectors, perpendicularity between a vector and a surface, and so forth.
We instruct the VLMs to first generate the constraints necessary for the first target pose, followed by the subsequent actions required after reaching that pose.
Fig. \ref{fig: method-bahavior} provides an illustrative example of this process.
Implementation details of this process are provided in the Appendix.

\noindent \textbf{Target Pose Planning.}
Upon obtaining manipulation constraints, we proceed to derive the sequence of post-grasp poses. 
This is equivalent to computing a sequence of SE(3) matrices such that, when applied to the parts of the object manipulated by the robotic arm, these parts satisfy the spatial geometric constraints. 
We operate under the assumption that the object part under manipulation and the robotic end-effector together constitute a rigid body. Consequently, these calculated SE(3) transformations can be directly applied to the robotic end-effector. 
We formalize the computation of the SE(3) matrix as a constrained optimization problem. Specifically, we compute a loss for each constraint, and then a nonlinear constraint solver is used to find the SE(3) matrix that minimizes the sum of these losses. 
Taking the constraint ``Vector 2 points downward'' from Fig.~\ref{fig: method-bahavior} as an example, the loss can be defined as the negative dot product of the normalized Vector 2 after SE(3) transformation and the vector $(0, 0, -1)$. 
After obtaining the first target pose, we solve for subsequent poses in alignment with the actions specified by VLMs. Concretely, we sequentially compute a new pose corresponding to each subsequent action. 
For example, for the action ``Move vertically down 7 cm,'' we simply subtract 7 cm from the current pose on the z-axis.
This process results in a complete set of post-grasp poses $\{P_1, P_2, ..., P_N\}$, with the transitions between adjacent poses facilitated by motion planning algorithms.
The detailed process for solving the SE(3) matrix and a comprehensive description of the subsequent actions can be found in the Appendix.

%% file: contents/experiments.tex
\begin{table*}[]
    \centering
    \begin{tabular}{l|ccccc}
         \toprule
         \textbf{Tasks}
         & \begin{tabular}[c]{@{}c@{}} \textbf{\method} \\ \textbf{(Ours)} \end{tabular}
         & \textbf{Voxposer}
         & \begin{tabular}[c]{@{}c@{}} \textbf{\method} \\ \textbf{w/o foundation} \end{tabular}
         & \begin{tabular}[c]{@{}c@{}} \textbf{\method} \\ \textbf{w/o coarse-to-fine} \end{tabular}
         & \begin{tabular}[c]{@{}c@{}} \textbf{\method} \\ \textbf{w/o constraint} \end{tabular} \\
         \midrule
         Hammer nail         & 30\% & 0\% & 0\% & 0\% & 10\%              \\
         Find scissors           & 70\% & 50\% & 10\% & 70\% & 70\%        \\
         Press button          & 80\%  & 10\%  & 10\% & 60\% & 20\%      \\
         Open drawer          & 80\%  & 40\%  & 10\% & 70\% & 30\%      \\
         Pour water      & 30\%  & 0\% & 0\% & 10\% & 0\%       \\
         Put eraser into drawer          & 80\% & 30\% & 30\% & 60\% & 80\%  \\
         Insert flower into vase             & 70\% & 0\% & 0\% & 60\% & 0\%  \\
         Put glasses onto shelf          & 60\% & 20\% & 30\% & 50\% & 60\% \\
         Put spoon into cup              & 60\% & 10\% & 0\%  & 30\% & 30\% \\
         Sweep nuts             & 70\% & 20\% & 20\% & 50\% & 70\% \\
         \midrule
         \textbf{Total}                       & \textbf{63\%} & 18\% & 11\% & 46\% & 37\% \\  
         \bottomrule
    \end{tabular}
    \caption{Quantitative results in real-world experiments. \method~successfully complete everyday manipulation tasks with a high success rate, demonstrating a profound physical understanding of scenes, significantly surpassing the baseline VoxPoser. Furthermore, we conduct ablation study to validate the importance of foundation models in our algorithm, as well as the design of coarse-to-fine grounding and constraint generation.}
    \label{tab:exps}
\end{table*}

\section{Experiments}
\label{sec:exp}

We first introduce the experimental setup in Section~\ref{sec:setup}.
Subsequently, we evaluate the performance of \method~in real-world manipulation tasks in Section~\ref{sec:our-results}.
Then we highlight \method's intriguing properties by comparing it with the baseline VoxPoser \cite{huang2023voxposer} in Section~\ref{sec:compare-voxposer}.
We further present an ablation study to analyze the contribution of key modules within our framework in Section~\ref{sec:ablation}.
Finally, we demonstrate that \method~can be seamlessly integrated with high-level task planning methods to accomplish complex long-horizon tasks in Section~\ref{sec:high-level}.

\subsection{Experimental Setup}
\label{sec:setup}

\noindent \textbf{Hardware.}
We set up a real-world tabletop environment. We use a Franka Emika Panda robot (a 7-DoF arm) and a 1-DoF parallel jaw gripper.
For perception, we mount two RGB-D cameras (Intel RealSense D435) at two opposite ends of the table and calibrate them.

\noindent \textbf{Tasks and Evaluations.}
We design 10 real-world manipulation tasks, each demanding a comprehensive understanding of the physical properties of objects.
See Fig. \ref{fig: method-overview} for illustrations of the tasks.
For each task, we evaluate all methods across 10 different variations of the environment, which encompass alterations in object types and their arrangements.
Detailed descriptions of the tasks are provided in the Appendix.

\noindent \textbf{VLMs and Prompting.}
We employ GPT-4V from~\href{https://openai.com/api/}{OpenAI API} as the VLM.
\method~involves minimal few-shot prompts to aid VLMs in comprehending their roles.
Additionally, the chain-of-thought technique~\cite{wei2022chain} is utilized to facilitate a deeper understanding of the scene by VLMs.
The full prompt is provided in the Appendix.

\noindent \textbf{Baselines.}
We compare with Voxposer~\cite{huang2023voxposer}, a method capable of synthesizing closed-loop robot trajectories without necessitating additional training through the utilization of a series of foundational models.
Following Huang et al \cite{huang2023voxposer}, we employ GPT-4 from~\href{https://openai.com/api/}{OpenAI API} as the LLM, and utilize the open-vocabulary detector Owl-ViT \cite{minderer2205simple} and Segment Anything \cite{kirillov2023segment} for perception.

\subsection{\method~for Real-World Manipulation}
\label{sec:our-results}
We study whether \method~can generate robot trajectories to perform real-world manipulation tasks.
The quantitative results are detailed in Table \ref{tab:exps}, while the Appendix showcases additional qualitative outcomes, including visualizations of part grounding results and manipulation constraints. 
We find that \method~achieves a remarkable success rate of 63\% across ten different tasks, significantly outperforming the VoxPoser baseline and various ablation variants (detailed in the following sections). 
A key factor in \method's superior performance is its leverage of common sense knowledge embedded in VLMs, which enables a fine-grained understanding of objects' physical properties during both part grounding and constraint generation phases. 
For example, in the part grounding phase, \method~accurately identifies the need to grasp the protective cover of an eraser in the \texttt{Put eraser on shelf} task, and recognizes the stem of the flower and the rim of the vase as critical parts in the \texttt{Insert flower into vase} task.
During the constraint generation phase, \method~comprehends that the spoon can be inserted vertically down into the cup in the \texttt{Put spoon into cup} task, and that the wooden stick needs to be aligned directly with the button in the \texttt{Press button} task.

\begin{figure}[t]
    \centering
    \includegraphics[width=0.5\textwidth]{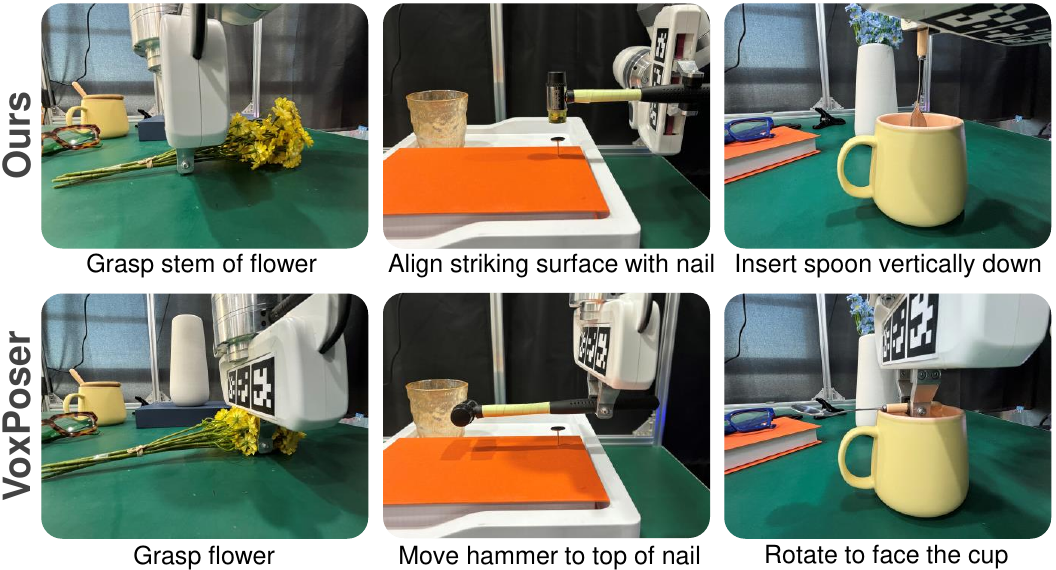}
    \caption{\textbf{Comparison with VoxPoser.} We illustrate the execution of \method~(top) and VoxPoser (bottom), demonstrating that \method~possesses a fine-grained physical understanding of scenes and can effectively handle rotation DoF. The tasks from left to right are sequentially \texttt{Insert flower into vase}, \texttt{Hammer nail}, \texttt{Put spoon into cup}. } 
    \label{fig: exp}
\end{figure}

\subsection{Understanding Properties of \method}
\label{sec:compare-voxposer}
In this section, we delve deeper into \method, shedding light on its intriguing properties through a comparative analysis with Voxposer, another method that utilizes the common sense knowledge embedded in foundation models to synthesize robot trajectories. \method~exhibits significant advantages in the following three aspects:

\noindent \textbf{Fine-Grained Physical Understanding.}
Many manipulation tasks require a nuanced physical understanding of the scene, which necessitates not only identifying object parts with fine granularity but also comprehending their intricate attributes. 
\method~excels in this aspect, employing a coarse-to-fine part grounding module to select grasping/task-relevant object parts, and then utilizing VLMs to provide their spatial geometry constraints.
In contrast, Voxposer only perceives objects in the scene as a whole. This coarse-grained level of comprehension often leads to failure in tasks that require precise operations.
For instance, in the \texttt{Insert flower into vase} task (shown in Fig. \ref{fig: exp} left), \method~ grasps the stem of the flower, whereas Voxposer seizes the petals.
In the \texttt{Hammer nail} task (shown in Fig. \ref{fig: exp} middle), \method~orients the hammer to align precisely with the nail, while Voxposer overlooks this fine-grained physical constraint, treating the hammer as a single rigid body.

\noindent \textbf{Simple Prompt Engineering.}
\method~demonstrates remarkable generalizability across a wide range of scenarios with minimal prompt engineering. 
In our \method~experiments, we employ just three examples to aid the VLMs in comprehending their roles. 
In contrast, Voxposer relies on highly complex prompts containing 85 hand-crafted examples. Its capability for reasoning predominantly stems from the provided prompts, thereby limiting its generalizability to new scenarios. When we attempt to simplify Voxposer's prompts, reducing the example count to three for each module, the system's performance drastically declines, resulting in almost complete failure across all evaluated tasks.

\noindent \textbf{Handling Rotation DoF.}
Robotic manipulation requires not just the movement of the end-effector to a specified location but also the precise control of its rotation. 
For example, in the \texttt{Pour water} task, it is essential to rotate the kettle to a certain angle to enable the water to flow out through the spout. 
\method~calculates the end-effector's 6-DoF pose by considering the spatial geometric constraints of key object parts within the scene, allowing for accurate and continuous control over rotation DoF. 
Conversely, Voxposer attempts to have LLMs directly specify the end-effector's rotation DoF based on simple examples in prompts, causing the output rotation values to be selected from a limited set of discrete options. This approach often overlooks the dynamic interactions and constraints between objects.
For example, in the \texttt{Put spoon into cup} (shown in Fig. \ref{fig: exp} right), \method~rotates the spoon to a vertical orientation, whereas Voxposer positions the robot's end-effector to face the cup, resulting in a collision between the spoon and the cup.

\subsection{Ablation Study}
\label{sec:ablation}
We next conduct a series of ablation studies to demonstrate the significance of the foundation model within our framework, as well as the design of coarse-to-fine grounding and constraint generation. The results are shown in Table~\ref{tab:exps}.
\subsubsection{\method~w/o foundation}
We eliminate the use of foundation vision-language models (GPT-4V).
Specifically, we substitute grasping/task-relevant parts grounding module with an open-vocabulary detector, Owl-ViT.
Additionally, we remove the constraint generation phase and instead compute post-grasp poses in a predefined rule-based manner (detailed in the Appendix).
The results, as presented in Table~\ref{tab:exps}, reveal that this approach encounters significant challenges, with a success rate of merely 11\% across all the tasks. This underscores the crucial role of the common sense knowledge embedded within VLMs. 
For example, in the \texttt{Sweep nuts} task, it becomes challenging to determine which tool in the scene is suitable for sweeping without the aid of VLMs.

\subsubsection{\method~w/o coarse-to-fine}
We eliminate the coarse-to-fine design in the grounding module, opting instead for direct utilization of fine-grained SoM and GPT-4V to select object parts within scenes.
Experimental results indicate that removing coarse-to-fine design leads to a performance decline, especially in tasks where identifying important parts accurately is challenging.
For example, in the \texttt{Hammer nail} tasks, the absence of the coarse-to-fine design makes this variant impossible to accurately identify the hammer's striking surface, leading to zero success rate for this task.
\subsubsection{\method~w/o constraint}
In this ablation study, we have the VLMs directly output numerical values for the post-grasp poses of the end-effector, instead of the constraints that need to be satisfied by the object being manipulated.
Experiments demonstrate that, for most manipulation tasks, directly deriving precise pose values from scene images is extremely challenging.
For instance, in the \texttt{Pour water} task, it's almost impossible for this variant to generate precise pose values to tilt the kettle to the correct pose.
In contrast, utilizing constraints given by VLMs to solve for post-grasp poses presents a more viable option.

\begin{figure*}[t]
    \centering
    \includegraphics[width=\textwidth]{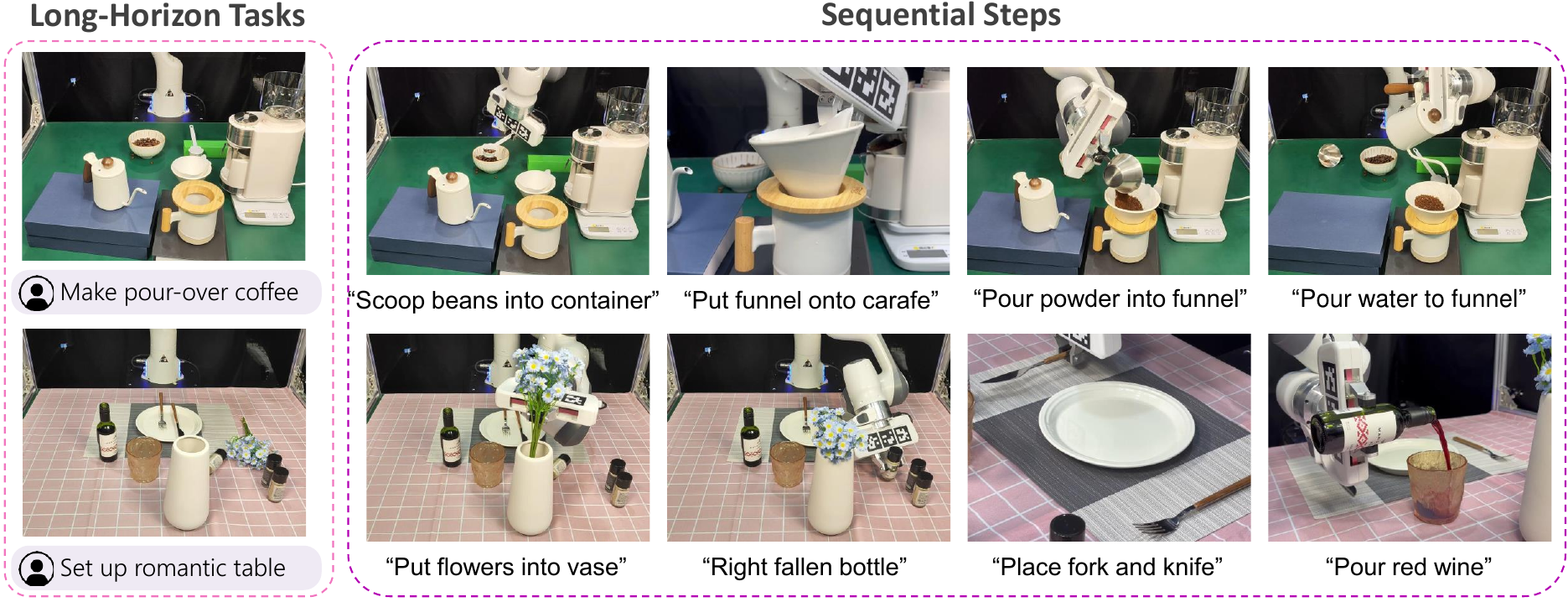}
    \caption{\textbf{Intergration with High-Level Planning.} We show the execution process of two long-horizon tasks: \texttt{Make pour-over coffee} and \texttt{Set up romantic table}. We demonstrate that \method~can be seamlessly integrated with high-level planning methods to accomplish complex long-horizon tasks.} 
    \label{fig: exp-high-level}
\end{figure*}

\subsection{Integration with High-Level Planning}
\label{sec:high-level}
High-level planning and low-level control are two critical and decoupled aspects of robotic task execution.
Our low-level control framework can be seamlessly integrated with high-level planning methods to accomplish complex long-horizon tasks.
We design two long-horizon tasks, \texttt{Make pour-over coffee} and \texttt{Set up romantic table}, to validate the effectiveness of this combination.
Not only do these two tasks need to be accurately decomposed into reasonable and actionable steps, but the execution of each step requires a profound understanding of the physical properties of the task-relevant objects.
Specifically, we employ \textsc{ViLa} \cite{hu2023look} as the high-level planning method to decompose the high-level instruction into a sequence of low-level control tasks.
Subsequently, these low-level control tasks are executed sequentially using \method.
Fig.~\ref{fig: exp-high-level} shows some environment rollouts.
Experiments demonstrate that \method, combined with high-level planning methods, can effectively complete long-horizon tasks, showcasing the potential of this combination for real-world applications.


%% file: contents/conclusions.tex
\section{DISCUSSION \& LIMITATIONS}
\label{sec:conclusion}
In this work, we present \method, a novel framework that leverages the common sense knowledge of foundation vision-language models to generate pose sequences for robotic manipulation tasks.
\method~operates effectively with simple prompt engineering without requiring any training.
Boasting a fine-grained physical understanding of scenes, \method~can generalize to open-world scenarios, handling open-set instructions and objects.
Moreover, \method~can be naturally combined with high-level planning algorithms to accomplish complex, long-horizon tasks.


\method~has a few limitations that future work can improve.
First, \method's capability to process complex objects is constrained by its reliance on simplistic geometric elements such as surfaces and vectors. This can be improved by incorporating more geometric elements into our modeling process.
Second, the VLMs currently in use are pre-trained on large-scale 2D images and lack a genuine grounding in the \textit{3D physical world}. This limitation hampers their ability to perform accurate spatial reasoning. Integrating 3D inputs, like point clouds, into the training phase of VLMs may alleviate this challenge.
Lastly, the existing VLMs produce only discrete textual outputs, whereas our framework essentially necessitates \textit{continuous} output values, like the coordinates of object parts. The development of foundation models that incorporate these capabilities remains a highly anticipated advancement.

%% file: contents/appendix.tex
\clearpage
\onecolumn
\appendix 

\subsection{Hardware Setup}
We set up a real-world tabletop environment. We use a Franka Emika Panda robot (a 7-DoF arm) and a 1-DoF parallel jaw gripper.
We use Franka ROS and MoveIt\footnote{\href{http://docs.ros.org/en/kinetic/api/moveit_tutorials/html/}{http://docs.ros.org/en/kinetic/api/moveit\_tutorials/html/}} to control the robot, which by default uses an RRT-Connect planner for motion planning.
For perception, we mount two RGB-D cameras (Intel RealSense D435) at two opposite ends (left and right from the top-down view) of the table and calibrate them.

\subsection{Tasks and Evaluations.}
We design 10 real-world manipulation tasks, each demanding a comprehensive understanding of the physical properties of objects.
We provide a detailed description of these tasks in Table~\ref{tab:task_details}.
For each task, we evaluate all methods across 10 different variations of the environment, which encompass alterations in object types and their arrangements.

\begin{table*}[h!]
    \small
    \centering
    \begin{tabular}{m{14em}m{40em}}

    \toprule
    \texttt{Hammer nail}
    & \textbf{Instruction:} ``Hammer the nail.'' \\
    & \textbf{Description:} 
    This task requires the robot to first grasp the handle of the hammer, then rotate it until its striking surface aligns with the surface of the nail, and finally hammer downwards. To accomplish this task, it is essential to accurately identify and model the hammer’s striking surface, handle and the nail’s surface. \\
    \noalign{\vskip 1mm}
    \hline
    \noalign{\vskip 1mm}
    \texttt{Find scissors}  & \textbf{Instruction:} ``Find scissors for me.'' \\
    & \textbf{Description:} In this task, the scissors may be partially obscured by other objects, such as books. The robot is required to first locate the scissors and then grasp its handle.\\
    
        \noalign{\vskip 1mm}
    \hline
    \noalign{\vskip 1mm}
    \texttt{Press button} & \textbf{Instruction:} ``Press the button with the stick.'' \\
    & \textbf{Description:} This task necessitates initially grasping the stick, then rotating it until its axis is directly aligned with the button, and finally pressing it. To accomplish this task, it is imperative to accurately identify and model the stick and the button. \\
    
        \noalign{\vskip 1mm}
    \hline
    \noalign{\vskip 1mm}
    \texttt{Open drawer}
     & \textbf{Instruction:} ``Open the drawer.'' \\
    & \textbf{Description:} This task requires the initial grasping of the drawer handle, followed by a linear pull along the handle's normal vector. \\
    
        \noalign{\vskip 1mm}
    \hline
    \noalign{\vskip 1mm}
    \texttt{Pour water} & \textbf{Instruction:} ``Pour water from kettle to funnel/cup.'' \\
    & \textbf{Description:} This task requires that the spout needs to be moved directly above the funnel, and the kettle needs to be rotated at a certain angle so that the water can flow out. This task imposes stringent demands on the robot's control over its rotation DoF.  \\

        \noalign{\vskip 1mm}
    \hline
    \noalign{\vskip 1mm}
    \texttt{Put eraser into drawer} & \textbf{Instruction:} ``Put eraser into the drawer.'' \\
    & \textbf{Description:}  In this task, a portion of the eraser is encapsulated by a protective cover, necessitating that the robot exclusively grasps this protective cover. \\

        \noalign{\vskip 1mm}
    \hline
    \noalign{\vskip 1mm}
    \texttt{Insert flower into vase } & \textbf{Instruction:} ``Put flowers into the vase.'' \\
    & \textbf{Description:}  This task requires first grasping the flower by its stem (not the petals), then moving the flower directly above the vase while rotating the flower to an upright position, and finally inserting it straight down into the vase. \\

        \noalign{\vskip 1mm}
    \hline
    \noalign{\vskip 1mm}
    \texttt{Put glasses onto shelf } & \textbf{Instruction:} ``Put glasses onto the shelf.'' \\
    & \textbf{Description:} 
    In this task, We need to utilize common sense knowledge to determine that, when picking up glasses, one should grasp the frame rather than the lenses.  \\
    
        \noalign{\vskip 1mm}
    \hline
    \noalign{\vskip 1mm}
    \texttt{Put spoon into cup } & \textbf{Instruction:} ``Put spoon into the cup.'' \\
    & \textbf{Description:}  This task requires first grasping the spoon's handle, then rotating it to the vertical direction, moving it directly above the cup, and finally inserting it vertically down into the cup. \\
    
        \noalign{\vskip 1mm}
    \hline
    \noalign{\vskip 1mm}
    \texttt{Sweep nuts } & \textbf{Instruction:} ``Select a tool to sweep nuts aside.'' \\
    & \textbf{Description:}  
    This task requires the robot to first identify a tool (e.g. rasp) suitable for sweeping nuts through common sense knowledge, and then to grasp the handle of the selected tool. \\
    
    \bottomrule
    
    \end{tabular}
    \caption{\textbf{A List of 10 Real-World Manipulation Tasks.} These tasks require a profound physical understanding of the scene. We provide the instructions used in our experiments and detailed descriptions for each task.}
    \label{tab:task_details}
\end{table*}

\subsection{VLMs and Prompting.}
We employ GPT-4V from~\href{https://openai.com/api/}{OpenAI API} as the VLM.
\method~involves minimal few-shot prompts to aid VLMs in comprehending their roles.
Additionally, the chain-of-thought technique~\cite{wei2022chain} is utilized to facilitate a deeper understanding of the scene by VLMs.
Prompts used in Section~\ref{grasp} and Section~\ref{behavior} can be found as follows:

\noindent \textbf{Coarse-Grained Grasping Object Grounding}:~\href{https://copa-2024.github.io/prompts/coarse_grained_grasping_object_grounding.pdf}{copa-2024.github.io/prompts/coarse\_grained\_grasping\_object\_grounding.pdf}  

\noindent \textbf{Fine-Grained Grasping Part Grounding}:~\href{https://copa-2024.github.io/prompts/fine_grained_grasping_part_grounding.pdf}{copa-2024.github.io/prompts/fine\_grained\_grasping\_part\_grounding.pdf}  

\noindent \textbf{Coarse-Grained Task-Relevant Object Grounding}:~\href{https://copa-2024.github.io/prompts/coarse_grained_relevant_object_grounding.pdf}{copa-2024.github.io/prompts/coarse\_grained\_relevant\_object\_grounding.pdf}  

\noindent \textbf{Fine-Grained Task-Relevant Part Grounding}:~\href{https://copa-2024.github.io/prompts/fine_grained_relevant_part_grounding.pdf}{copa-2024.github.io/prompts/fine\_grained\_relevant\_part\_grounding.pdf} 

\noindent \textbf{Constraint Generation}:~\href{https://copa-2024.github.io/prompts/constraint_generation.pdf}{copa-2024.github.io/prompts/constraint\_generation.pdf} 

\subsection{Baselines.}
We compare with Voxposer~\cite{huang2023voxposer}, a method capable of synthesizing closed-loop robot trajectories without necessitating additional training through the utilization of a series of foundational models.
Following Huang et al \cite{huang2023voxposer}, we employ GPT-4 from~\href{https://openai.com/api/}{OpenAI API} as the LLM, and utilize the open-vocabulary detector Owl-ViT \cite{minderer2205simple} and Segment Anything \cite{kirillov2023segment} for perception.
Additionally, we adopt their real-world prompt as the prompt for Voxposer in our experiments.


\subsection{Robotic Arm Filtering.}
Based on the camera's extrinsic parameters, we render the robot's URDF model onto the camera plane, thereby obtaining the robot's mask.

\subsection{Part Modeling and Annotation.}
In the task-aware motion planning phase (Section \ref{behavior}), we need to model task-relevant parts selected by the grounding module and annotate them in the scene image. We complete this through the following stages:

\noindent \textbf{Part Modeling.}
We model identified task-relevant parts as vectors or surfaces.
Specifically, we commence by obtaining the minimum bounding rectangle for each part.
Parts with an aspect ratio exceeding a predetermined threshold are considered slender and are modeled as vectors.
The remaining parts are modeled as surfaces. 

\noindent \textbf{Part Formulation.}
We need to obtain the mathematical representation of each part on the 2D image in this stage.
For parts modeled as vectors, we first perform linear regression to fit a line that best corresponds to their 2D masks, then identify the intersection points of the line with the boundaries of the parts, which serve as the endpoints of the vector.
For parts modeled as surfaces, we first employ the RANSAC algorithm to determine their 3D center points and normal vectors, which are then projected onto the 2D image.

\noindent \textbf{Part Annotation.}
Now we need to annotate the parts according to their formulation on the scene image.
First, each part is masked with color and translucency on the image.
Then for the parts modeled as vectors, we connect the two endpoints and put a numerical label adjacent to the endpoint farther from the robot arm.
For parts modeled as faces, we mark the center point and normal vector, and label adjacent to the center point.

\subsection{Details of Constraints.}
For each task, we obtain a set of constraints through vision-language models.
We then utilize optimization algorithms (e.g., the BFGS algorithm or Trust-Region Constrained Optimization) to solve for an SE(3) matrix that minimizes the cumulative loss associated with these constraints.
In Table~\ref{tab:constraint}, we provide a detailed description of the constraints used in our experiments, along with their corresponding loss calculation methods.
In the descriptions of these constraints, \texttt{Point A} and \texttt{Vector A} are located on the object being manipulated, and thus require an SE(3) transformation when calculating loss.
Other points and vectors are considered static and do not require SE(3) transformation.

We define the SE(3) matrix we need to solve as follows:
\begin{equation}
\begin{aligned}
\mathbf{T} &=
\begin{bmatrix}
    \mathbf{R} & \mathbf{t}\\
    \mathbf{0^T} & 1
\end{bmatrix} \in \mathbb{SE}(3),
\end{aligned}
\end{equation}
where Euclidean group $\mathbb{SE}(3):=\{\mathbf{R},\mathbf{t}\ |\ \mathbf{R}\in \mathbb{SO}(3), \mathbf{t}\in \mathbb{R}^3\}$.
We denote the points as $p \in \mathbb{R}^3$, and the normalized vectors as $V \in \mathbb{R}^3$. Furthermore, we denote $\mathcal{T}$ as the SE(3) transformation, which can be applied to both points and vectors:
\begin{equation}
\begin{aligned}
    \mathcal{T}(p) = \mathbf{R}p + \mathbf{t}, \quad
    \mathcal{T}(V) = \mathbf{R}V,
    \label{eq:trans}
\end{aligned}
\end{equation}

\begin{table*}[h!]
    \small
    \centering
    \begin{tabular}{m{20em}m{30em}}

    \toprule
    \textbf{Descriptions of Constraints} & \textbf{Loss Calculation} \\
    
        \noalign{\vskip 1mm}
    \hline
    \noalign{\vskip 1mm}
    \texttt{Vector A} and \texttt{Vector B} are on the same line, with the opposite direction.  & 
    $loss = \left \| \mathcal{T}(V_A) \times V_B \right \| + \left \| (\mathcal{T}(p_A) - p_B) \times V_B \right \| +  \left \| \mathcal{T}(V_A) + V_B \right \| $ \\
    
        \noalign{\vskip 1mm}
    \hline
    \noalign{\vskip 1mm}
    The target position of \texttt{Point A} is $x$ cm along \texttt{Vector B} from \texttt{Point C}'s current position. & $loss = \left | (\mathcal{T}(p_A) - p_C) \cdot V_{B} - x \right | + \left \| (\mathcal{T}(p_A) - p_C) \times V_{B} \right \|$  \\
    
        \noalign{\vskip 1mm}
    \hline
    \noalign{\vskip 1mm}
    \texttt{Vector A} is parallel to the table surface.
     & $loss = \left | \mathcal{T}(V_A) \cdot V_{table} \right | $ \\
    
        \noalign{\vskip 1mm}
    \hline
    \noalign{\vskip 1mm}

    \texttt{Point A} is $x$ cm above the table surface.
     & $loss = \left | (\mathcal{T}(p_A) - p_{table}) \cdot V_{table} - x \right | $ \\
    
        \noalign{\vskip 1mm}
    \hline
    \noalign{\vskip 1mm}
    
    \texttt{Vector A} is perpendicular to the table surface. & $loss = \left \| \mathcal{T}(V_A) \times V_{table} \right \|$  \\
    
    \bottomrule
    
    \end{tabular}
    \caption{\textbf{Desciptions of Constraints and Their Corresponding Loss Calculation Methods.} $\left \| \cdot  \right \|$ represents $l_2$-norm. The variables $p_{table}$ and $V_{table}$ respectively denote the coordinate and the normal vector of the table, with their values being $(0.5, 0, 0.07)$ and $(0, 0, 1)$ respectively.
    Each \texttt{Vector} corresponds to a \texttt{Point}. The normal vector of the part modeled as the surface corresponds to the center point of the surface, while the point corresponding to the part modeled as the vector is the point farther away from the robotic arm among its two endpoints.}
    \label{tab:constraint}
\end{table*}

\subsection{Details of Subsequent Actions.}
In Table \ref{tab:sub_act}, we provide a detailed description of the subsequent actions utilized in our experiments, along with their corresponding methodologies for calculating new poses.

\begin{table*}[h!]
    \small
    \centering
    \begin{tabular}{m{18em}m{25em}}

    \toprule
    \textbf{Descriptions of Subsequent Actions} & \textbf{New Pose Calculation} \\
    
        \noalign{\vskip 1mm}
    \hline
    \noalign{\vskip 1mm}
    Move vertically down $x$ cm.  & 
    Subtract $x$ cm from the current pose on the z-axis. \\
    
        \noalign{\vskip 1mm}
    \hline
    \noalign{\vskip 1mm}
    Move forward $x$ cm. &Move $x$ cm along the current orientation of the end-effector.  \\
    
        \noalign{\vskip 1mm}
    \hline
    \noalign{\vskip 1mm}

    Open the gripper.
     & Open the gripper. \\
    
        \noalign{\vskip 1mm}
    \hline
    \noalign{\vskip 1mm}
    
    End-effector rotates 180 degrees. & Rotate the joint corresponding to the end-effector (the 7th joint for Franka Emika Panda robot) by 180 degrees.  \\
    
    \bottomrule
    
    \end{tabular}
    \caption{\textbf{Desciptions of subsequent actions and their corresponding new pose calculation methods.} }
    \label{tab:sub_act}
\end{table*}

\subsection{Predefined Rule-Based Post-Grasp Pose Generation.}
We design a rule-based method to replace the constraint generation module within our framework to generate post-grasp poses.
This method entails a prescribed pose calculation protocol specific to each format 
of instruction. The formats of the instructions along with their corresponding post-grasp poses calculation methodologies 
are detailed in Table~\ref{tab:rule-pose}.

\begin{table*}[h!]
    \small
    \centering
    \begin{tabular}{m{12em}m{16em}}

    \toprule
    \textbf{Instruction Format} & \textbf{Post-Grasp Pose Calculation} \\
    
        \noalign{\vskip 1mm}
    \hline
    \noalign{\vskip 1mm}
    Hammer \texttt{A}.  & \begin{tabular}[l]{@{}l@{}} 1. Move hammer to 5 cm above \texttt{A}. \\ 2. Move vertically down 6 cm. \end{tabular}
    \\ 
    
        \noalign{\vskip 1mm}
    \hline
    \noalign{\vskip 1mm}
    Press \texttt{A} with \texttt{B}. & \begin{tabular}[l]{@{}l@{}} 1. Move \texttt{B} to 5 cm above \texttt{A}. \\ 2. Move vertically down 6 cm. \end{tabular}  \\
    
        \noalign{\vskip 1mm}
    \hline
    \noalign{\vskip 1mm}
    Open \texttt{A}.
     & 1. Move backward 10 cm. \\
    
        \noalign{\vskip 1mm}
    \hline
    \noalign{\vskip 1mm}

    Pour water from \texttt{A} to \texttt{B}.
     & \begin{tabular}[l]{@{}l@{}} 1. Move \texttt{A} to 5 cm above \texttt{B}. \\ 2. End-effector rotates 180 degrees. \end{tabular} \\
    
        \noalign{\vskip 1mm}
    \hline
    \noalign{\vskip 1mm}
    
    Put \texttt{A} into \texttt{B}. & \begin{tabular}[l]{@{}l@{}} 1. Move \texttt{A} to 5 cm above \texttt{B}. \\ 2. Open the gripper. \end{tabular}  \\
    
    \bottomrule
    
    \end{tabular}
    \caption{\textbf{Predefined rule-based pose calculation methods.} \texttt{A} and \texttt{B} in the instruction can refer to any object.}
    \label{tab:rule-pose}
\end{table*}

\subsection{More Visualization.}
\begin{figure}[t]
    \centering
    \includegraphics[width=\textwidth]{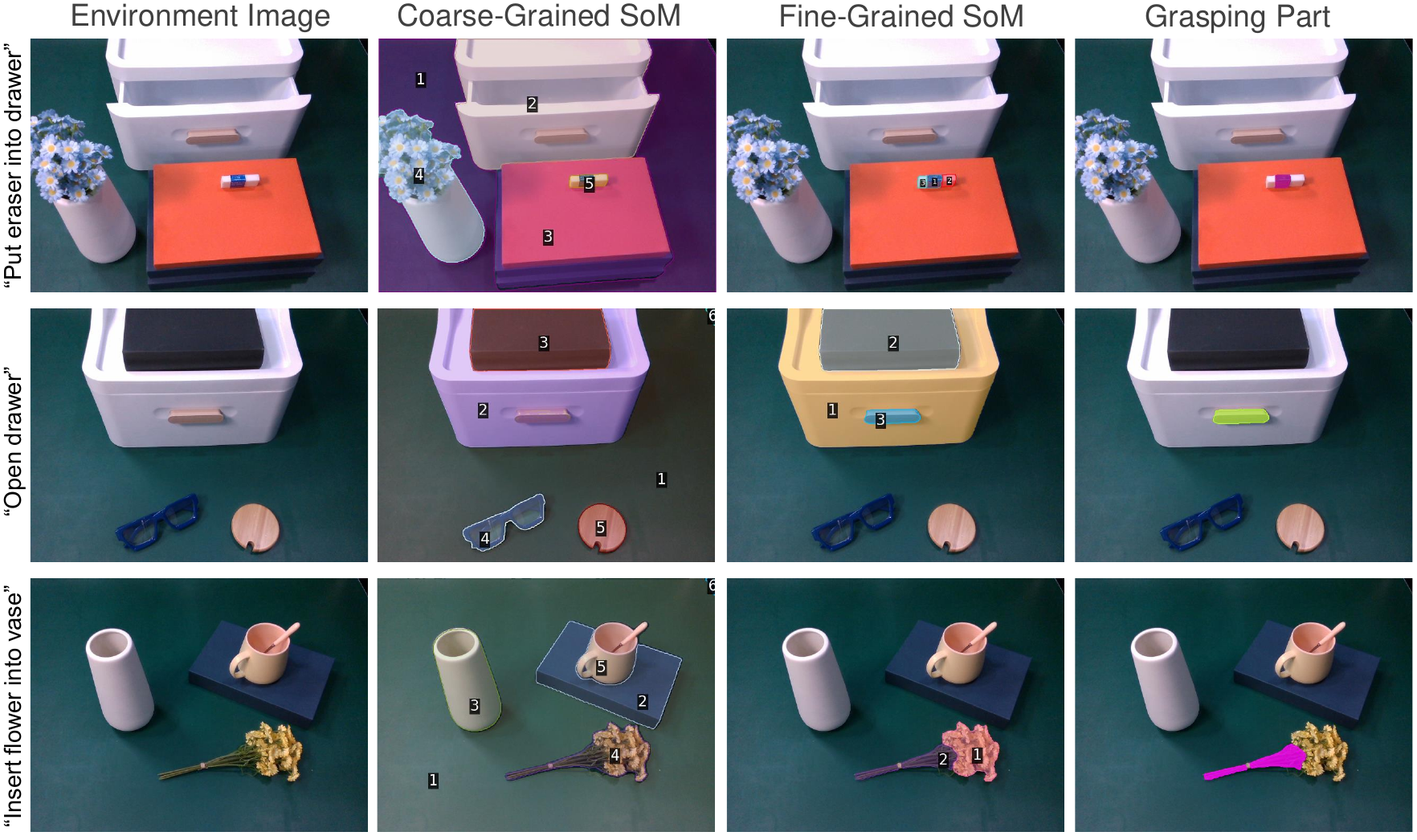}
    \caption{\textbf{Visualization for \textcolor{mygreen}{Grounding Module}.} } 
    \label{fig: more-grounding}
\end{figure}

\begin{figure}[t]
    \centering
    \includegraphics[width=\textwidth]{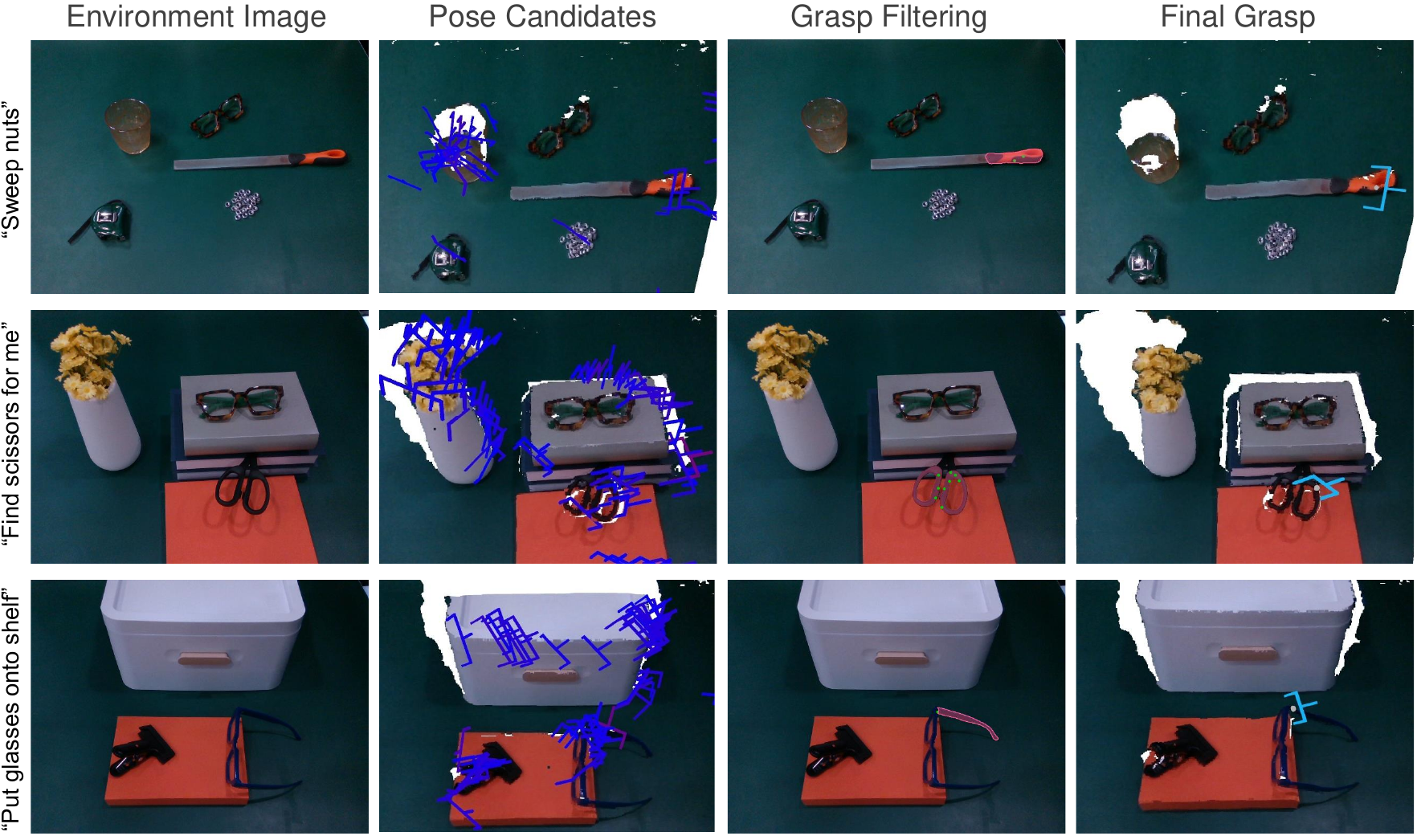}
    \caption{\textbf{Visualization for \textcolor{myorange}{Task-Oriented Grasping}}. } 
    \label{fig: more-grasping}
\end{figure}

\begin{figure}[t]
    \centering
    \includegraphics[width=\textwidth]{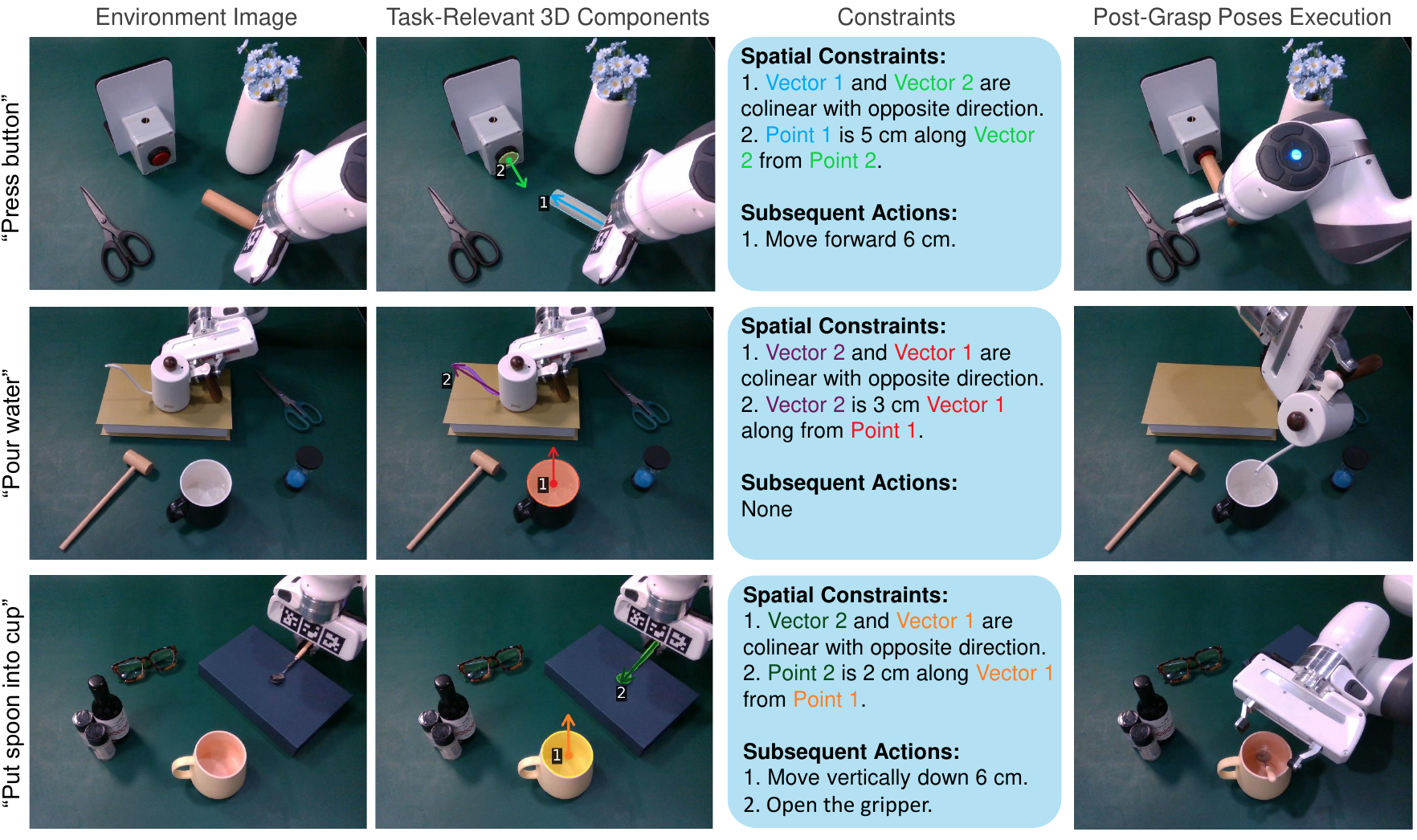}
    \caption{\textbf{Visualization for \textcolor{myblue}{Task-Aware Motion Planning}.} } 
    \label{fig: more-behavior}
\end{figure}

Additional visualization for \textcolor{mygreen}{grounding module} are presented in Fig. \ref{fig: more-grounding}, for \textcolor{myorange}{task-oriented grasping} in Fig. \ref{fig: more-grasping}, and for \textcolor{myblue}{task-relevant motion planning} in Fig. \ref{fig: more-behavior}.

